\pdfoutput=1

\documentclass[11pt]{article}

\usepackage[]{acl}

\usepackage{times}
\usepackage{latexsym}
\usepackage{algorithm}
\usepackage{algorithmic}
\usepackage{CJKutf8}
\usepackage{subfigure}
\usepackage{adjustbox}
\usepackage{multirow}
\usepackage{booktabs}
\usepackage{tabularx}

\newcommand{\para}[1]{\vspace{.05in}\noindent\textbf{#1}}
\usepackage[T1]{fontenc}

\usepackage[utf8]{inputenc}

\usepackage{microtype}

\newcommand{\hust}{$^1$}
\newcommand{\alibaba}{$^2$}
\newcommand{\scut}{$^3$}
\newcommand{\ku}{$^4$}

\title{Geo-Encoder: A Chunk-Argument Bi-Encoder Framework for \\ Chinese Geographic Re-Ranking}

\author{Yong Cao\hust, Ruixue Ding\alibaba, Boli Chen\alibaba, Xianzhi Li\hust\footnotemark[2], Min Chen\scut, \\ {\bf Daniel Hershcovich\ku, Pengjun Xie\alibaba, and Fei Huang\alibaba} \\
{\hust}Huazhong University of Science and Technology 
{\alibaba}Alibaba Group, Hangzhou, China \\
{\scut}School of Computer Science and Engineering, South China University of Technology \\
{\ku}Department of Computer Science, University of Copenhagen \\
\normalsize{\texttt{\{yongcao\_epic,xzli\}@hust.edu.cn, minchen@ieee.org, dh@di.ku.dk}} \\ \normalsize{\texttt{\{ada.drx, boli.cbl, chengchen.xpj, f.huang\}@alibaba-inc.com}}}

\begin{document}
\maketitle
\begin{abstract}
Chinese geographic re-ranking task aims to find the most relevant addresses among retrieved candidates, which is crucial for location-related services such as navigation maps. Unlike the general sentences, geographic contexts are closely intertwined with geographical concepts, from general spans (e.g., province) to specific spans (e.g., road). Given this feature, we propose an innovative framework, namely \textit{Geo-Encoder}, to more effectively integrate Chinese geographical semantics into re-ranking pipelines. Our methodology begins by employing off-the-shelf tools to associate text with geographical spans, treating them as chunking units. Then, we present a multi-task learning module to simultaneously acquire an effective attention matrix that determines chunk contributions to extra semantic representations. Furthermore, we put forth an asynchronous update mechanism for the proposed addition task, aiming to guide the model capable of effectively focusing on specific chunks. Experiments on two distinct Chinese geographic re-ranking datasets, show that the \textit{Geo-Encoder} achieves significant improvements when compared to state-of-the-art baselines. Notably, it leads to a substantial improvement in the Hit@1 score of MGEO-BERT, increasing it by 6.22\% from 62.76 to 68.98 on the GeoTES dataset.
\end{abstract}

\renewcommand{\thefootnote}{\fnsymbol{footnote}} 
\footnotetext[2]{Corresponding author.}
\renewcommand{\thefootnote}{\arabic{footnote}} 

\section{Introduction}

\begin{figure}[t]
\centering
  \includegraphics[width=1.0\columnwidth]{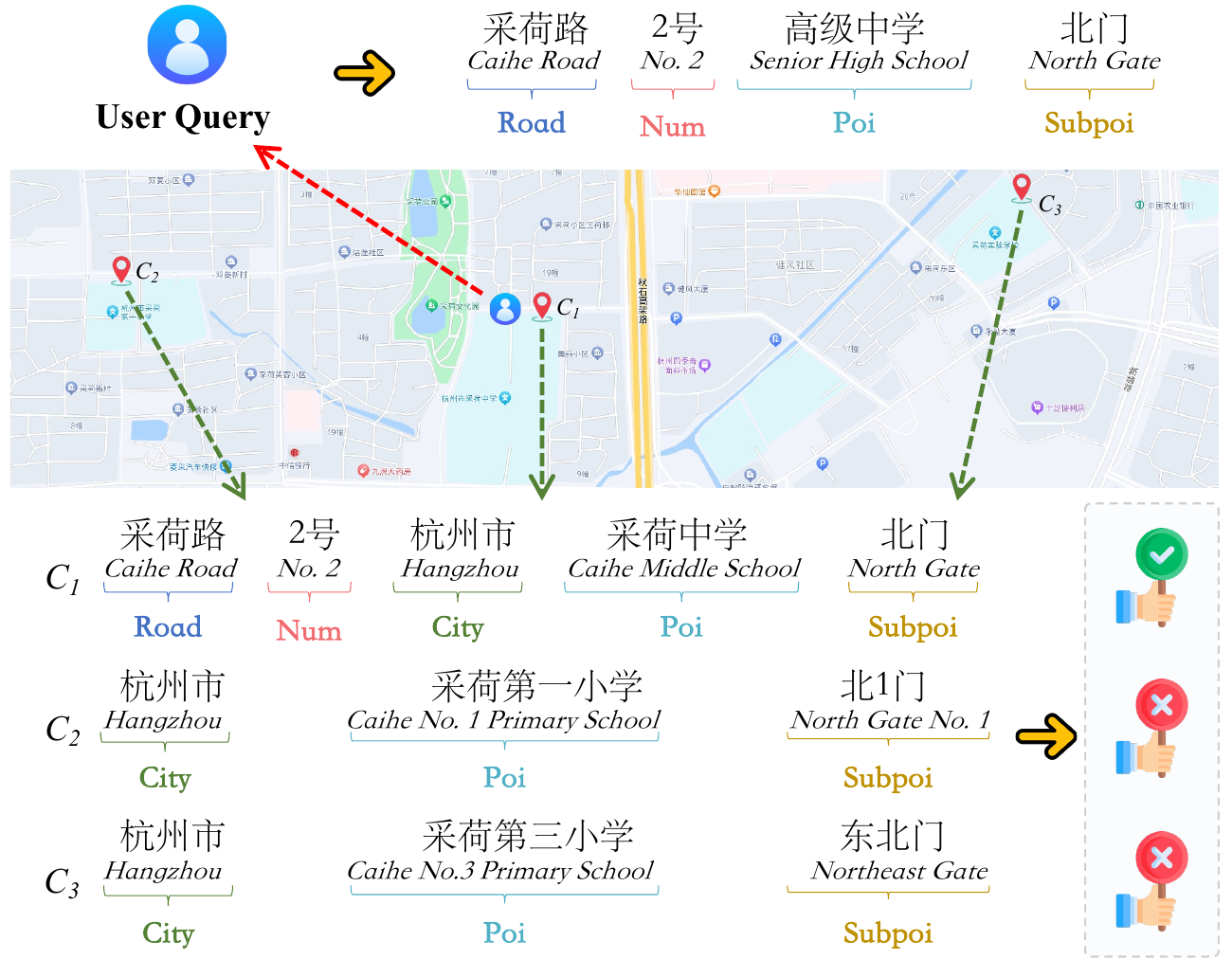}
  \caption{Overview of the Chinese Geographic re-ranking task. The process begins with the user query being subjected to word chunking, segmenting it into meaningful units. Lastly, \textit{Geo-Encoder} is employed to enhance semantic representation and re-ranking.}
  \label{fig:figure1}
\end{figure}

Chinese geographic re-ranking (CGR) is a sub-task of semantic matching, aiming to identify the most relevant geographic context towards given queries and retrieved candidates~\cite{zhao2019incorporating, MacAvaney2020, yates2021pretrained}. It is a crucial task that serves many downstream applications such as navigation maps (e.g., Gaode Maps), autonomous driving (e.g., Tesla), E-commerce system (e.g., Taobao), etc.~\cite{jia2017dynamic, avvenuti2018gsp}. Unlike general query expressions, Chinese geographic sentences exhibit a distinct attribute in their linear-chain structural semantics \cite{li-etal-2019-neural-chinese}. This peculiarity arises from the fact that Chinese addresses often comprise distinct meaningful address segments, termed as geographic chunks in linguistic terms~\cite{abney1991parsing}. These chunks adhere to a specific format, organizing from the general (e.g., province) to the more specific (e.g., road). For example, as is shown in Figure \ref{fig:figure1}, given a Chinese address \begin{CJK}{UTF8}{gbsn}``\emph{采荷路2号高级中学北门 (North Gate of Caihe Road No.2 Senior High School)}''\end{CJK}, we can deconstruct it into several such chunks: \begin{CJK}{UTF8}{gbsn}``\emph{采荷路 (Caihe Road)}''\end{CJK}, \begin{CJK}{UTF8}{gbsn}``\emph{2号 (No.2)}''\end{CJK}, \begin{CJK}{UTF8}{gbsn}``\emph{高级中学 (Senior High School)}''\end{CJK}, \begin{CJK}{UTF8}{gbsn}``\emph{北门 (North Gate)}''\end{CJK}.

Conventional approaches~\cite{reimers2019sentence, humeau2019poly, khattab2020colbert} addressing the CGR task often directly employ pre-trained language models (PLMs) to encode given geographic texts into embeddings, which are subsequently subjected to re-ranking through similarity calculation techniques like cosine or euclidean distance measures. Recent works~\cite{yuan2020spatio, huang2022ernie, ding2023multi} in this field extend beyond mere geographic context utilization and encompass an expansive range of data sources, including point-of-interest information, multi-modal data, and user behavioral attributes \cite{liu-etal-2021-geo-bert, hofmann2022geographic, huang2022ernie} with a larger neural model. The outcome of this integration is characterized by notable enhancements, achieved through the fusion of external geographic knowledge. Furthermore, cutting-edge domain-adaptation frameworks have been introduced to facilitate effective fusion of multi-domain data, such as PALM \cite{zhao2019incorporating}, STDGAT \cite{yuan2020spatio}, etc.

However, despite the effectiveness of existing attempts in leveraging geographic knowledge, these methods failed to fully harness the intrinsic potential of the geographic context itself. Therefore, in this paper, we aim to shift our focus towards the geographic context by exploiting its distinctive linear-chain attributes. To achieve this, we employ off-the-shelf tools (e.g. MGEO tagging\footnote{\url{https://modelscope.cn/models/damo/mgeo_geographic_elements_tagging_chinese_base/summary}.} and part-of-speech (POS) \footnote{POS tagging is based on jieba: \url{https://github.com/fxsjy/jieba}.} for the approximate annotation of each geographic text with pertinent geographic chunks. For example, as illustrated in Figure \ref{fig:figure1}, we annotate the text \begin{CJK}{UTF8}{gbsn}``\emph{采荷路 (Caihe Road)}''\end{CJK} with the label \textit{Road}, \begin{CJK}{UTF8}{gbsn}``\emph{2号 (No.2)}''\end{CJK} with \textit{Num}, etc.

\textbf{Firstly}, building upon this foundation, we introduce an additional task that revolves learning the similarity between different components of these annotated chunks. This involves the formulation of an attention matrix, which governs the contributions of these chunks to the semantic representations. Our motivation is that general chunks tend to be less diverse across queries and candidates, and specific chunks possess a higher degree of distinctiveness. \textbf{Secondly}, we put forth a novel asynchronous update speed mechanism for the attention matrix. This mechanism is designed to empower the model to effectively focus its attention on the more specific chunks, thereby enhancing its discernment capabilities. \textbf{Lastly}, we advocate for the integration of the pure bi-encoder approach during the inference period. This strategy ensures a harmonious balance between performance and computational efficiency, safeguarding the efficacy of the model in both academic and industrial scenarios.

In summary, our key contributions are as follows: 1) We introduce a multi-task learning framework, denoted as \textit{Geo-Encoder}, which serves as a pioneering approach to integrate component similarity; 2) We present an asynchronous update mechanism, to distinguish specific chunks effectively; 3) Except evaluation on benchmark dataset, we collect and publish a nationwide geographic dataset in China, named GeoIND. Experimental results on two distinct Chinese geographic re-ranking datasets demonstrate the superiority of our \textit{Geo-Encoder} over competitive methods.  Our code and datasets are available at: \url{https://github.com/yongcaoplus/CGR_damo}.



\section{Related Work}

\paragraph{Semantic Matching and Re-Ranking.} Semantic matching is a widely-concerned task in natural language processing, including retrieval and re-ranking process~\cite{zhao2019incorporating, yates2021pretrained}. Different from retrieval task, re-ranking generally deal with smaller candidates. 
Within this domain, researchers employ bi-encoders to encode given queries and candidates separately by using the shared parameters, such as ESIM~\cite{chen2017enhanced}, SBERT~\cite{reimers2019sentence}, ColBERT~\cite{khattab2020colbert}, etc. And after the emergence of pre-trained models, such as RoBERTa~\cite{liu2019roberta}, ERNIE \cite{sun2021ernie}, cross-encoders were proposed to jointly encode text and promote the information interaction~\cite{humeau2019poly, nie2020dc, ye2022fast}. Besides, to better represent sentences, external knowledge and late interactions were widely explored. For example,~\citet{xia2021using} utilized a word similarity matrix to assign term weights for given tokens, and ~\citet{peng-etal-2022-predicate} introduced predicate-argument spans to enhance representation. Notably, the bi-encoder is industry-preferred for its efficiency thus we adopted it in our paper.

\paragraph{Chinese Geographic Text Representation.} Most existing approaches focused on encoding geographic text by external knowledge in two aspects: (1) position data, such as PALM~\cite{zhao2019incorporating}, encoding positional relationship of query and candidates, STDGAT~\cite{yuan2020spatio}, considering Spatio-temporal features, etc.; (2) geographic knowledge, such as GeoL~\cite{huang2022ernie}, using knowledge related to user behaviors, and MGeo~\cite{ding2023multi}, proposing using multi-modal dataset. However, the geographic text encoding method among the above approaches is not well-explored. Besides, parsing Chinese geographic text into chunks is also a key technical issue~\cite{li-etal-2019-neural-chinese}. Generally, address parsing is quite similar to Chinese word segmentation. Existing attempts to Chinese word segmentation includes CRF models~\cite{zhao2006effective}, latent-variable variants~\cite{sun2009discriminative}, neural transition-based segmentation method~\cite{zhang-etal-2016-transition-based}, and chart-based models~\cite{stern-etal-2017-minimal, kitaev-klein-2018-constituency}, etc. However, while these models benefit from external geographic knowledge, exploring text representation optimization beyond direct PTMs encoding is still crucial.

\section{Our Approach}

\subsection{Task Definition and Overview}
In Chinese Geographic Re-ranking (CGR) task setting, the available dataset $\{X\}$ is formed as query-candidate pairs. Let Q denotes queries and C as retrieved candidates, where C is the corresponding candidates list of each query from Q. Both Q and C are composed of \textit{l} separated tokens, where \{X\} = $\{X \in (Q, C) | X = x_1, x_2, ..., x_l\}$. 
The objective of CGR is to model the highest possibility of C. Thus, the bi-encoder framework, depicted in Figure \ref{fig:network}(a), can be formalized as:

\begin{equation}
    c = \mathop{\arg\max}_C r_\theta \left(f_\theta(Q), f_\theta(C)\right)
\end{equation}
where $f_\theta$ denotes encoding function (we adopt PLMs here), to encode given text into vectors, $c$ ($\in C$) is the model output and $r_\theta$ denotes similarity evaluation function, such as dot multiple and cosine similarity, to assign a similarity score for each candidate. Also, the cross-encoder framework, depicted in Figure \ref{fig:network}(b) can be formalized as: 

\begin{equation}
    c = \mathop{\arg\max}_C r_\theta \left(f_\theta(Q, C)\right)
\end{equation}

\begin{figure}[t]
\centering
  \includegraphics[width=1.0\columnwidth]{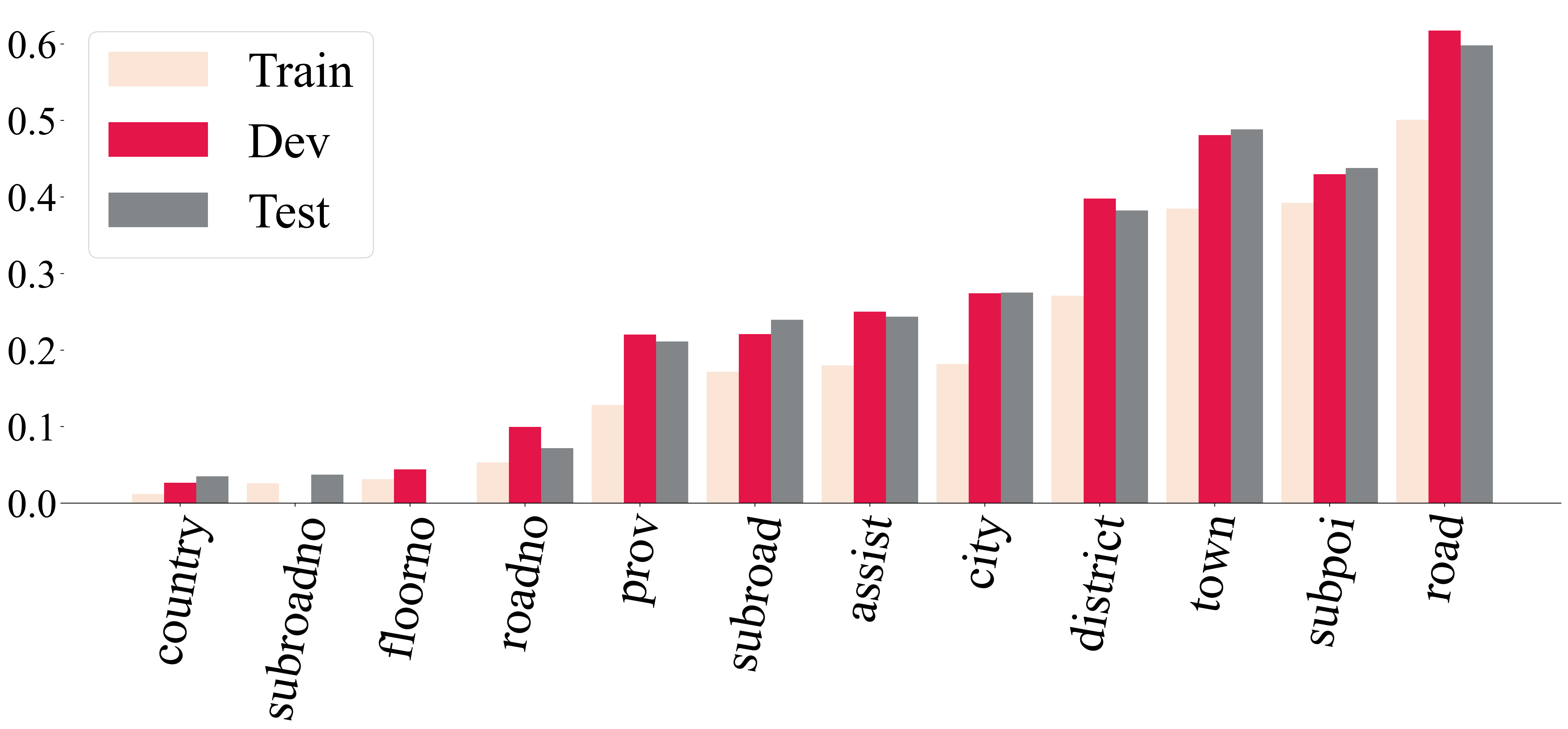}
  \caption{The information entropy of~\citet{li-etal-2019-neural-chinese}, indicate that specific chunks (e.g., \emph{road}) exhibit greater diversity compared to general ones (e.g., \emph{country}).}
  \label{fig:chunk_distribution}
\end{figure}

\begin{figure*}[t]
\centering
  \includegraphics[width=0.98\textwidth]{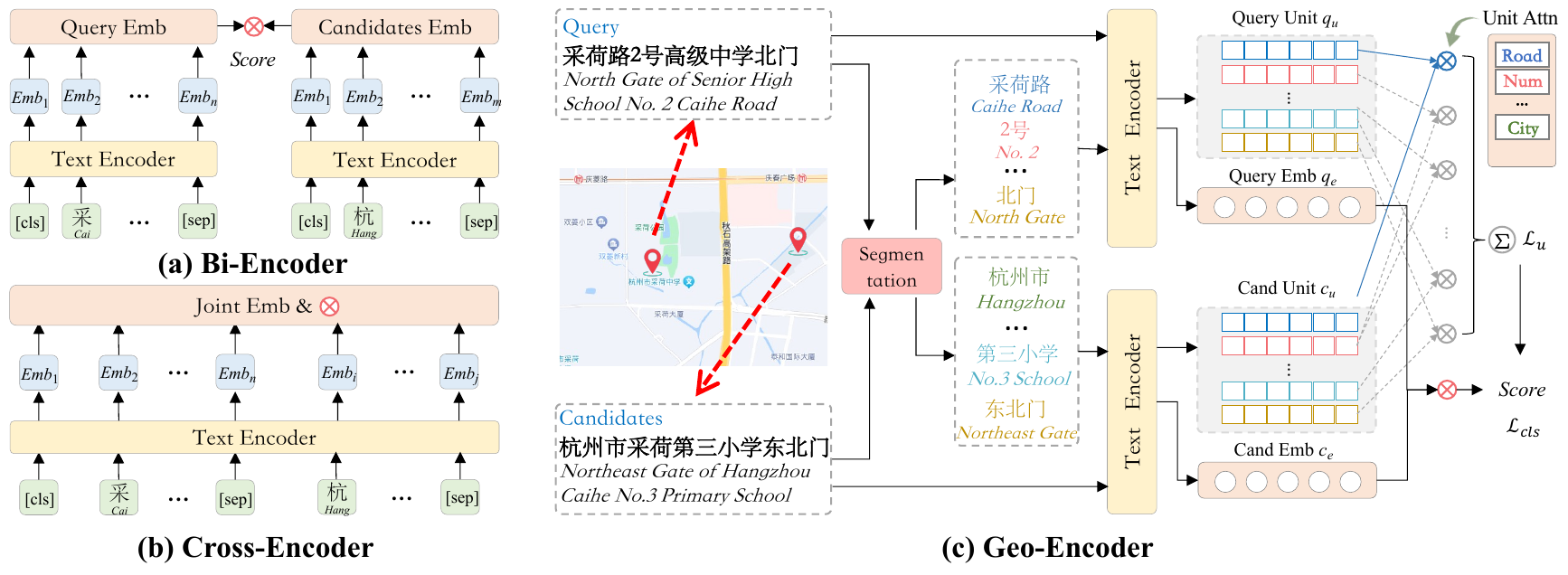}
  \caption{Architecture of re-ranking models and our proposed Geo-Encoder. The left shows the bi-encoder and cross-encoder models, and the right shows our proposed Geo-Encoder, which parsing geographic text into chunking units and jointly encoding with global features and unit attention mechanism. $\otimes$ denotes similarity calculation.}
  \label{fig:network}
\end{figure*}

Most current attempts directly deploy PTMs to encode geographic texts into embeddings \cite{yuan2020spatio, huang2022ernie, ding2023multi}, ignoring the linear-chain structure characteristic of geographic text. To quantify this distinction, we calculate the entropy score of geographic chunking datasets from \cite{li-etal-2019-neural-chinese} as shown in Figure \ref{fig:chunk_distribution}. Obviously, the specific chunks (e.g. road, town, etc.) hold a higher entropy score among all sets, revealing more diversity than the general chunks (e.g. country, province, etc.). Therefore, it can be further inferred that specific chunk components contribute unequally to the semantic representation of sentences, indicating that specific chunks play a more substantial role than general ones.

In our approach, we strive to enhance the encoding process through a two-step strategy. Firstly, we segment the provided geographic text into chunks and introduce a novel approach to learn both the attention matrix governing chunk contributions and component semantic representation as an additional task. Secondly, we introduce an asynchronous update mechanism for the attention matrix and model parameters. This mechanism is aimed at enabling the model to efficiently acquire the skill of focusing on specific chunks. Finally, we present our training and inference details. The detailed framework of our proposed method, called \textit{Geo-Encoder}, is shown in Figure \ref{fig:network}(c).

\subsection{Geographic Chunking}

Chinese addresses typically consist of multiple meaningful address segments, often referred to as "geographic chunks"~\cite{abney1991parsing}. These addresses follow a structured pattern, progressing hierarchically from the general (e.g., province) to specific ones (e.g., road)~\cite{li-etal-2019-neural-chinese}. In contrast to conventional Chinese segmentation methods, geographic chunking demands tools of heightened sensitivity tailored to geographical units. These tools necessitate fine-tuning using dedicated Chinese address corpora. Consequently, we adopt the MGEO tagging tool to facilitate the acquisition of precise geographic annotations for our benchmark datasets \cite{DBLP:journals/corr/abs-2210-10293, 2206.12608, ding2023multi}. 

MGEO stands as a pre-trained model with multi-modal datasets, encompassing both geographic context and points of interest. It is designed to cater to various downstream tasks, including geographic entity alignment and address element tagging, among others. In our current framework, however, we exclusively leverage MGEO to provide chunk annotations, without employing it for the purpose of encoding contextual information. 
Then, dataset $\{X\}$ is extended as $\{X_u\} =\{x \in (Q, C, Q_u, C_u) | X = x_1, x_2, ..., x_n\} $, where $Q_u$ and $C_u$ denotes query and candidates chunking units. For example, given a Chinese address \begin{CJK}{UTF8}{gbsn}``\emph{南京市新城科技园3栋5单元(Unit \#5, Building \#3, Sci-Tech Park, Nanjing City.)}''\end{CJK}, we can parse them by MGEO into: \begin{CJK}{UTF8}{gbsn}``\emph{南京市(Nanjing City)}''\end{CJK} -- city, \begin{CJK}{UTF8}{gbsn}``\emph{新城科技园(Sci-Tech Park)}''\end{CJK} -- devzone, \begin{CJK}{UTF8}{gbsn}``\emph{3栋(Building \#3)}''\end{CJK} -- houseno, \begin{CJK}{UTF8}{gbsn}``\emph{5单元(Unit \#5)}''\end{CJK} -- cellno.

\subsection{Chunking Contribution Learning}
Utilizing the chunked dataset denoted as $\{X_u\}$, we proceed to employ a pre-trained language model for the encoding process. This yields the representations of \textit{[CLS]} embedding $e_{cls}^q$ and token embedding $e_{1:l}^q$ from geographic text:

\begin{equation}
    e_{cls}^q, e_{1:l}^q =Encoder(q), q \in Q 
\end{equation}
where $Encoder$ denotes PTMs. And correspondingly we can get candidates features $e_{cls}^c$ and $e_{1:l}^c$. Given chunking annotations, we initialize a zeros query component embeddings $\{ U^Q| u^q_i \in U^Q \}, i = \{ 1,2, \cdots, M \}$ and we can further update query component embeddings $u_{i}^q$ by:

\begin{equation}
    u_{i}^q = mean(\Gamma (e_{1:l}^q, I_i^q))
    \label{eq:4}
\end{equation}
where $\Gamma(\cdot)$ is the Index function to obtain component token embeddings, M is the total amount of chunk categories, and $I_i^q$ is the index number acquired by the tokenizer of the $Encoder$ from the chunk's location to the corresponding query. Similarly, component embeddings $\{U^C| u^c_i \in U^C\}$ can also be obtained. We can also get candidates' component embeddings $u_{i}^c$ similar with Eq. \ref{eq:4}.

To incorporate token-level embeddings, the ColBERT model \cite{khattab2020colbert} introduced a multi-attention mechanism, which facilitates subsequent interactions between queries and candidates. This technique has demonstrated improved efficacy in re-ranking tasks. Nonetheless, it is essential to acknowledge that the ColBERT method entails significant additional computational resources. In light of this, our work introduces an innovative multi-task learning module that incorporates only geographic chunking component embeddings and utilizes an attention matrix to fuse results. This approach is designed to address the need for efficient resource utilization while maintaining or potentially improving performance. 

Specifically, we define an attention matrix that can be learned along the training process, denoted as $W_U$. Then, we can get the predictions from component embeddings:

\begin{equation}
    Score_{u} = (U^Q * W^U) * (U^C * W^U)
\end{equation}
 We use dot multiplication to obtain the similarity scores of given queries and candidates. Thus, for components embeddings, we  can obtain the component similarity loss $\mathcal{L}_{u}$ as:

\begin{equation}
    \mathcal{L}_{u} = \Phi (Score_{u}, Y)
\end{equation}
where $Y$ represents the ground truth ranking results, and $\Phi(\cdot)$ signifies the cross-entropy loss function.

As for the primary task, we use \textit{[CLS]} representation as sentence encoded features, and we can obtain the semantic similarity loss $\mathcal{L}_{cls}$ as:
\begin{equation}
    \mathcal{L}_{cls} = \Phi (E_{cls}^Q * E_{cls}^C, Y)
\end{equation}
where $e_{cls}^q \in E_{cls}^Q$ and $e_{cls}^c \in E_{cls}^C$. 

\paragraph{Discussion.} Due to the components of each geographic text being quite different, introducing feature concatenation strategy in CGR task is not reasonable. Therefore, we proposed to use an universal component embeddings for queries $U^Q$ and candidates $U^C$, and initialize them as zero matrices. It follows that empty components would yield no contributions to the final representations. Similarly, components that do not align appropriately between the queries and candidates would also have no impact.

\subsection{Asynchronous Update Mechanism}
For multi-task learning, a common concern is the disparate challenges faced by models when learning multiple tasks simultaneously, often leading to variations in convergence rates \cite{lu2017fully, he2017adaptively}. In our pursuit to tackle this quandary within our designated task, we deviate from established methodologies seen in prior literature \cite{isonuma-etal-2017-extractive, hashimoto-etal-2017-joint, nishino-etal-2019-keeping, pfeiffer-etal-2020-mad}. Instead, we propose an innovative approach involving the integration of an asynchronous update mechanism, which allocates enhanced focus on training steps pertaining to distinct tasks. To formalize our proposition, the update of parameter $w_u (w_u \in W_U)$ is as:

\begin{equation}
    w_u’ = w_u + \lambda \cdot \nabla w_u \cdot \gamma
\end{equation}
where $\gamma$ is a hyper-parameter to adjust training speed, which can set by grid search or empirically. 

\paragraph{Discussion.} Our insights is that the fast distinction of specific geographic chunks should conceivably be more amenable and expedited for the model's learning process. Consequently, the matrix $W_U$ could feasibly adapt to more substantial increments in learning steps compared to those attributed to language model parameters.

\subsection{Training and Inference}

During the training process of CGR, we deploy our proposed framework \textit{Geo-Encoder} of Figure \ref{fig:network}(c). The model can be optimized by jointly minimizing the semantic similarity loss and component similarity loss: 

\begin{equation}
    \mathcal{L} = \mathcal{L}_{cls} + \mathcal{L}_{u}
\end{equation}

During the inference phase, a notable concern arises from the time-intensive nature of indexing and calculating component embeddings, particularly when extrapolated to scenarios involving an extensive pool of candidates. To circumvent this challenge, we directly adopt a bi-encoder framework for conducting inference, as visually depicted in Figure \ref{fig:network}(a). 

\paragraph{Discussion.} Our rationale for introducing components stems from a deliberate consideration of the trade-off between training and inference aspects. The underlying objective is to facilitate the model in exhibiting a heightened sensitivity towards specific chunks as opposed to general ones. This endeavor has yielded demonstrably effective outcomes in our experimental evaluations. Conversely, during the inference phase, we eliminate the necessity for component predictions, thereby leading to a marked improvement in computational efficiency. This assertion will be substantiated in the subsequent section.

\section{Experiment}

\subsection{Datasets}
\label{ax:dataset_description}
To comprehensively validate the efficacy of our Geo-Encoder, we 
prepared two representative
Chinese geographical datasets: (i) \textbf{GeoTES}: a widely-recognized, large-scale benchmark dataset, and (ii) \textbf{GeoIND}: our collected moderately-sized, real-world industry dataset. The statistical details concerning the two datasets are presented in Table \ref{tb:dataset_statistics}.

\para{Geographic Textual Similarity Benchmark (GeoTES):} 
This large-scale dataset comprises queries meticulously crafted by human annotators and was amassed within the location of Hangzhou, China.\footnote{The dataset can be downloaded here: \url{https://modelscope.cn/datasets/damo/GeoGLUE/summary}.} The dataset's meticulous annotation was executed by a panel of 20 participants and four domain experts. Encompassing a total of 90,000 queries, each complemented by 20/40 retrieved candidates, this dataset extends its scope beyond geographical text, encapsulating supplementary point of interests (POIs) data. Please refer to Appendix for more details.

\para{Industry Geographic dataset (GeoIND):} 
For a broader validation, we re-organize and format an additional real-world dataset named GeoIndustry, sourced from a geographic search engine. This dataset underwent rigorous cleaning and filtration procedures, effectively eliminating noise and erroneous queries. In contrast to GeoTES, this dataset exhibits an intermediary scale, yet it boasts a substantial geographical coverage. 
We will make it publicly available upon the publication of our work.

\subsection{Baselines}
To assess the effectiveness of our Geo-Encoder, we undertake a comprehensive comparative analysis via representative bi-encoder methodologies. It's pertinent to mention that our assessment confines itself exclusively to geographic text data, with the exclusion of Points of Interest (POIs) or other modal data. Our selected baselines include:

\begin{itemize}
    \item \textbf{Word2Vec} \cite{mikolov2013efficient}. A traditional method captured semantic relationships between words and encoded words as dense vector embeddings.\footnote{Reproduced by text2vec package\cite{Text2vec}: \url{https://github.com/shibing624/text2vec}.}
    \item \textbf{Glove} (Pennington et al. 2014). It encapsulated both global and local semantic information and served for contextual understanding.
    \item \textbf{SBERT} \cite{reimers2019sentence}. A popular bi-encoder model that can effectively and efficiently serve for re-ranking task.\footnote{https://github.com/UKPLab/sentence-transformers.}
    \item \textbf{Argument-Encoder} \cite{peng-etal-2022-predicate}. It first proposed that concatenate predicate-argument embedding as extra representations can enhance re-ranking task.\footnote{We reproduce this method by replacing the predicate-argument with specific geographic-argument.} 
    \item \textbf{MGEO} \cite{ding2023multi}. By applying geographic POIs information to fuse external knowledge into encoder, this method achieves state-of-the-art results in current task.\footnote{We compare three backbone models with MGEO in text-only modal data, including BERT \cite{devlin-etal-2019-bert}, RoBERTa \cite{liu2019roberta}, and ERNIE 3.0 \cite{sun2021ernie}.}
\end{itemize}

Importantly, in real-world scenarios, accounting for computational efficiency is imperative. Therefore, in light of this consideration, we opt for the bi-encoder approach coupled with the current backbone models, rather than adopting the cross-encoder methodology or large language models. 

\begin{table}[t]
\resizebox{0.48\textwidth}{!}{
	\centering
	\begin{tabular}{c|cccccc}
	\toprule
	\textbf{Benchmark} & \textbf{Sets} & \textbf{Query}  & \textbf{Tokens} & \textbf{ASL} & \textbf{Cands} \\ 
        \midrule 
	\multirow{3}*{GeoTES} & Train & 50,000  & 3,599 & 18.8 & 20 \\
	& Dev & 20,000  & 3,322 & 17.2 & 40 \\
	& Test & 20,000  & 3,351 & 17.1 & 40 \\ \midrule 
	\multirow{3}*{GeoIND} & Train & 7,359  & 3,768 & 15.1 & 20 \\
	& Dev & 2,453  & 3,376 & 15.1 & 20 \\
	& Test & 2,469  & 2,900 & 15.0 & 20 \\
	\bottomrule
\end{tabular}}
\caption{\label{tb:dataset_statistics} The statistics of two datasets. \textit{Tokens} denotes vocabularies counts, \textit{ASL} denotes the average sentence length, and \textit{Cands} represents candidates numbers.}
\end{table}

\begin{table*}[t]
\centering
\resizebox{1.0\textwidth}{!}{
\begin{tabular}{l|cccc|cccc}
\toprule
\multicolumn{1}{l|}{\multirow{2}{*}{Model}} & \multicolumn{4}{c|}{\textbf{GeoTES}}      & \multicolumn{4}{c}{\textbf{GeoIND}}    \\
\multicolumn{1}{c|}{}                          & Hit@1 & Hit@3 & NDCG@1 & MRR@3  & Hit@1 & Hit@3  & NDCG@1 & MRR@3 \\ \midrule 
Word2vec \cite{mikolov2013efficient} & 19.26 & 30.60 & 28.79 & 24.15 & 47.79 & 71.69 & 66.15 & 58.27\\
Glove (Pennington et al. 2014) & 48.02 & 67.33 & 63.32 & 59.35 & 52.38 & 74.87 & 71.95 & 69.35 \\
SBERT \cite{reimers2019sentence} & 24.22 & 51.22 & 46.65 & 35.80 & 42.20 & 71.24 & 64.56 & 54.92 \\
Argument-Encoder \cite{peng-etal-2022-predicate} &   56.54  &  80.01  &  73.47  &  67.08 &  59.58  & 85.54   &  78.61  &  71.19  \\
MGEO-BERT \cite{ding2023multi}    & 62.76 & 80.89 & 75.95 & 70.87
             & 64.12 & 88.66 & 81.35 & 75.04 \\
\emph{\textbf{Geo-Encoder}} &  \textbf{68.98}    &   \textbf{85.82}   & \textbf{81.11}   & \textbf{76.56}  &  \textbf{66.71}    &  \textbf{89.35}   &  \textbf{82.78}   &  \textbf{76.99}    \\ \midrule
MGEO-ERNIE \cite{ding2023multi}      & 67.50 & 84.54 & 79.60 & 75.15
            & 63.95 & 87.89 & 81.06 & 74.60 \\
\emph{\textbf{Geo-Encoder}}   &  \textbf{68.66}  &  \textbf{85.64}   &  \textbf{80.75}   &  \textbf{76.30} &   \textbf{65.33}  &    \textbf{89.06}   &   \textbf{82.10}    &   \textbf{75.98}   \\ \midrule
MGEO-RoBERTa \cite{ding2023multi}     &  68.74 & 85.16 & 80.63 & 76.15 
                & 63.63 & 88.70 & 81.62 & 74.81 \\
\emph{\textbf{Geo-Encoder}} & \textbf{70.39}   &  \textbf{86.69}  &  \textbf{81.97}    & \textbf{77.72}  &  \textbf{67.27}    &   \textbf{90.28}     &  \textbf{83.61}    &   \textbf{77.56}   \\ \bottomrule
\end{tabular}}
\caption{\label{main_results} Main results on GeoTES and GeoIND, where bold values indicate the best performance within each column. Our proposed method consistently outperforms all three baselines across all metrics on both datasets.}
\end{table*}

\subsection{Experimental Setting}
\paragraph{Evaluation Metrics.} Following previous re-ranking tasks \cite{qu-etal-2021-rocketqa, ding2023multi}, we use Hit@K(K=1,3), NDCG@1 \cite{jarvelin2002cumulated} and MRR@3 to evaluate the performance across all models. Specifically, Hit@K quantifies the proportion of retrieved candidates that include at least one correct item within the top K ranks. NDCG@1 is a graded relevance measure that takes into account the positions of relevant items in the ranked list. MRR@3 calculates the average of the reciprocal ranks of the top-3 correct answers in the ranked list. 

\paragraph{Hyper-parameters.} For finetuing, we set the learning rate is set as 1e-5 for RoBERTa and 5e-5 for BERT and ERNIE. We finetune models for 50 epochs with early stopping after 3 epochs of no improvement in Hit@1 on the validation set. We conduct our experiment on a single A100 GPU and optimize all the models with Adam optimizer, where the batch size is set to 32. And followed by \citet{ding2023multi}, we decrease the embedding dimension from 768 to 256.

\subsection{Main Results}
We have conducted a rigorous comparison between our method with the aforementioned baselines and the results are presented in Table~\ref{main_results}.

Firstly, it is evident that our proposed approach achieves a remarkable state-of-the-art performance across all evaluated metrics, surpassing the performance exhibited by all alternative methods. This observation provides compelling evidence that our Geo-Encoder yields significant enhancements over multiple baseline models. Particularly, our method improves the Hit@1 score of BERT by 6.62\% from 62.76 to 68.98 on GeoTES dataset, and by 2.59\% from 64.12 to 66.71 on GeoIND dataset. 

Secondly, comparing three different backbone pre-trained models, RoBERTa performs emerges as the superior candidate, surpassing both BERT and ERNIE. This advantage can be attributed to RoBERTa's augmented network depth and its exposure to a comprehensive training corpus, endowing it with a heightened capacity for contextual comprehension and modeling than other models.

Thirdly, a notable trend is that the GeoTES dataset is marginally more amenable to learning compared to the GeoIND dataset, a phenomenon primarily attributed to its significantly larger scale, which is 6.76 times greater. This distinction is corroborated by the highest attained Hit@1 score of 70.39 on the GeoTES dataset, as opposed to the score of 67.27 observed on the GeoIND dataset.

Furthermore, we can also conclude that conventional encoding methodologies such as word2vec, GloVe, and SBERT exhibit subpar performance in CGR tasks. And it is pertinent to mention that in the context of the CGR task, cosine similarity tends to exhibit suboptimal performance compared to dot multiplication. This is evident from the fact that SBERT yields lower performance scores across both datasets. Similarly, the argument-enhancement techniques and the MGEO bi-encoder manifest a consistently underwhelming performance across both datasets.

\section{Analysis and Discussion}
In this section, we first conduct a comprehensive analysis of our proposed modules, and then discuss the advantages of using Geo chunking for CGR task. Lastly, a detailed exploration of hyper-parameter setting and the learned chunking attention metric is presented for a deeper understanding.

\subsection{Fix Contribution vs. Learning Weight}

In accordance with human experiential knowledge, the common practice involves the gradual differentiation of an address by sequentially hypothesizing the constitutive chunking elements, transitioning from general segments to more precise ones. Evidently, the generalized segments found among the pool of candidates tend to exhibit significant similarity, thus warranting a diminished influence on the semantic alignment process towards given queries. On the basis of this underlying hypothesis, we have formulated a comparative experiment intended to investigate the potential benefits arising from the dynamic allocation of chunk contributions in the context of representation learning.

\begin{table}[]
\begin{adjustbox}{width=0.45\textwidth,center}
\begin{tabular}{l|cccc}
\toprule
Method       & Hit@1 & Hit@3 & NDCG@1 & MRR@3   \\
\midrule
\multicolumn{5}{c}{GeoTES}                  \\
\midrule
baseline & 62.76 & 80.89 & 75.95 & 70.87 \\
\quad \textit{w} Fixed\_1.0    & 68.08 & 85.35 & 80.48 & 75.84 \\
\quad \textit{w} Fixed\_0.5  &  66.02    &   83.91   &  78.97  & 74.03  \\ 
\quad \textit{w} Fixed\_0.1  &   68.19  &  84.95    &  80.31   &  75.70   \\ 
\quad \textit{w} POS \textit{\textbf{(Ours)}} &   68.25   & 85.55  &  80.65 & 76.02  \\
\quad \textit{w} Geo \textit{\textbf{(Ours)}} & \textbf{68.98}    &   \textbf{85.82}   &  \textbf{81.11}   & \textbf{76.56}  \\
\midrule
\multicolumn{5}{c}{GeoIND}                  \\
\midrule
baseline & 64.12 & 88.66 & 81.35 & 75.04 \\
\quad \textit{w} Fixed\_1.0    & 65.61 & 89.59 & 82.47 & 76.39 \\
\quad \textit{w} Fixed\_0.5  &  65.69     &   89.06   &  82.28   &  76.23      \\
\quad \textit{w} Fixed\_0.1  &   64.20    &   87.85   & 81.14  &  74.77   \\  
\quad \textit{w} POS \textit{\textbf{(Ours)}} &   65.21    &  89.59  &  82.24 & 76.06  \\  
\quad \textit{w} Geo \textit{\textbf{(Ours)}} & \textbf{66.71}    &    \textbf{89.35}   &  \textbf{82.78}   &  \textbf{76.99} \\
\bottomrule
\end{tabular}
\end{adjustbox}
\caption{\label{tb:ablation_study} Ablation study on GeoTES and GeoIndust Sets, including exclude automatic attention update mechanism and geographic chunking information.}
\end{table}

Specifically, we constructed an experimental framework wherein the dynamic interplay of chunk contributions is examined. This is realized by configuring the attention matrices within the \textit{Geo-Encoder} architecture as constant values, effectively precluding gradient updates. We fix the attention weight with the values of 0.1, 0.5, and 1.0 respectively, thereby probing the impact of different attention allocation strategies on the learning process.

As is shown in Table \ref{tb:ablation_study}, we can find that the imposition of fixed attention matrices contributes to a reduction in the performance of the \textit{Geo-Encoder} across both datasets. Besides, the diverse initialization schemes for these attention matrices yield distinct effects among datasets. Within the GeoTES dataset, an initialization ratio of 0.1 yields optimal results, indicating a higher reliance on the sentence-level \textit{[CLS]} representation. Conversely, the GeoIND dataset attains peak performance when the ratio is set to 1.0, implying a contrasting attention distribution trend. Lastly, we find that even exclude the automatic update of attention matrices, the resultant performance still surpasses that of the baseline models. This outcome underscores the benefits derived from the incorporation of chunking information, substantiating its constructive impact on enhancing the overall model performance.

\subsection{Geo Chunking vs. General Chunking}

Subsequently, our investigation delves deeper into the influence of geographic chunks (Geo) by conducting a substitution experiment wherein these chunks are replaced with Part-of-Speech (POS) tagging results. To achieve this, we employ the jieba POS tagging tool to restructure the two datasets\footnote{To ensure a fair comparison, we manually select relevant POS labels (e.g., quantity, noun, position, etc.), while excluding irrelevant ones (e.g., tone, punctuation, preposition, etc.). Further details can be found in the Appendix.}. It is essential to note that the core distinction between POS and Geo lies in the target of segmentation: while GEO is geared towards geographic ontology, POS is more focused on semantic components.

The results, as depicted in Table \ref{tb:ablation_study}, yield an interesting observation that employing POS tagging can benefit both datasets, signified by the obvious superior performance of POS when compared to the baseline. This favorable outcome can be attributed to the additional representation and multi-task learning introduced by our approach. Nevertheless, it is noteworthy that despite the advantageous performance of POS, it lags behind Geo in terms of performance. This discrepancy further underscores the pivotal role played by geographic chunks in the context of the CGR task. Irrespective of the approach used for segmentation, our framework consistently exhibits better performance, thereby reinforcing \textit{Geo-Encoder}'s adaptability and efficacy. Therefore, our proposed framework transcends the confines of the Chinese task, and holds relevance and applicability to other languages or tasks characterized by sentence structures that align with linear-chain attributes.

\begin{table}[t]
\begin{adjustbox}{width=0.46\textwidth,center}
\begin{tabular}{l|cccc}
\toprule
\multirow{2}{*}{Method} & \multicolumn{2}{c}{GeoTES}    & \multicolumn{2}{c}{GeoIND}    \\
 &
  \begin{tabular}[c]{@{}l@{}}Training\\ (hour)\end{tabular} &
  \begin{tabular}[c]{@{}l@{}}Inference\\ (ms/case)\end{tabular} &
  \begin{tabular}[c]{@{}l@{}}Training\\ (hour)\end{tabular} &
  \begin{tabular}[c]{@{}l@{}}Inference\\ (ms/case)\end{tabular} \\
  \midrule
Word2vec                & \multicolumn{1}{c}{--} & \phantom{0}5.9  & \multicolumn{1}{c}{--} & \phantom{0}3.5  \\
Augment-Encoder         & 6.24                   & 32.7 & 1.52                   & 15.8 \\
MEGO-BERT             & 4.50                   & 33.8 & 0.92                   & 18.9 \\
\emph{\textbf{Geo-Encoder}} & 5.94                   & 35.6 & 1.25                   & 19.5 \\
\bottomrule
\end{tabular}
\end{adjustbox}
\caption{\label{tb:effency_proof} The statistics of training and inference time across different bi-encoder baseline models and our proposed \textit{Geo-Encoder} on GeoTES and GeoIND datasets.}
\end{table}

\begin{figure}[t]
    \centering
    \includegraphics[width=0.45\textwidth]{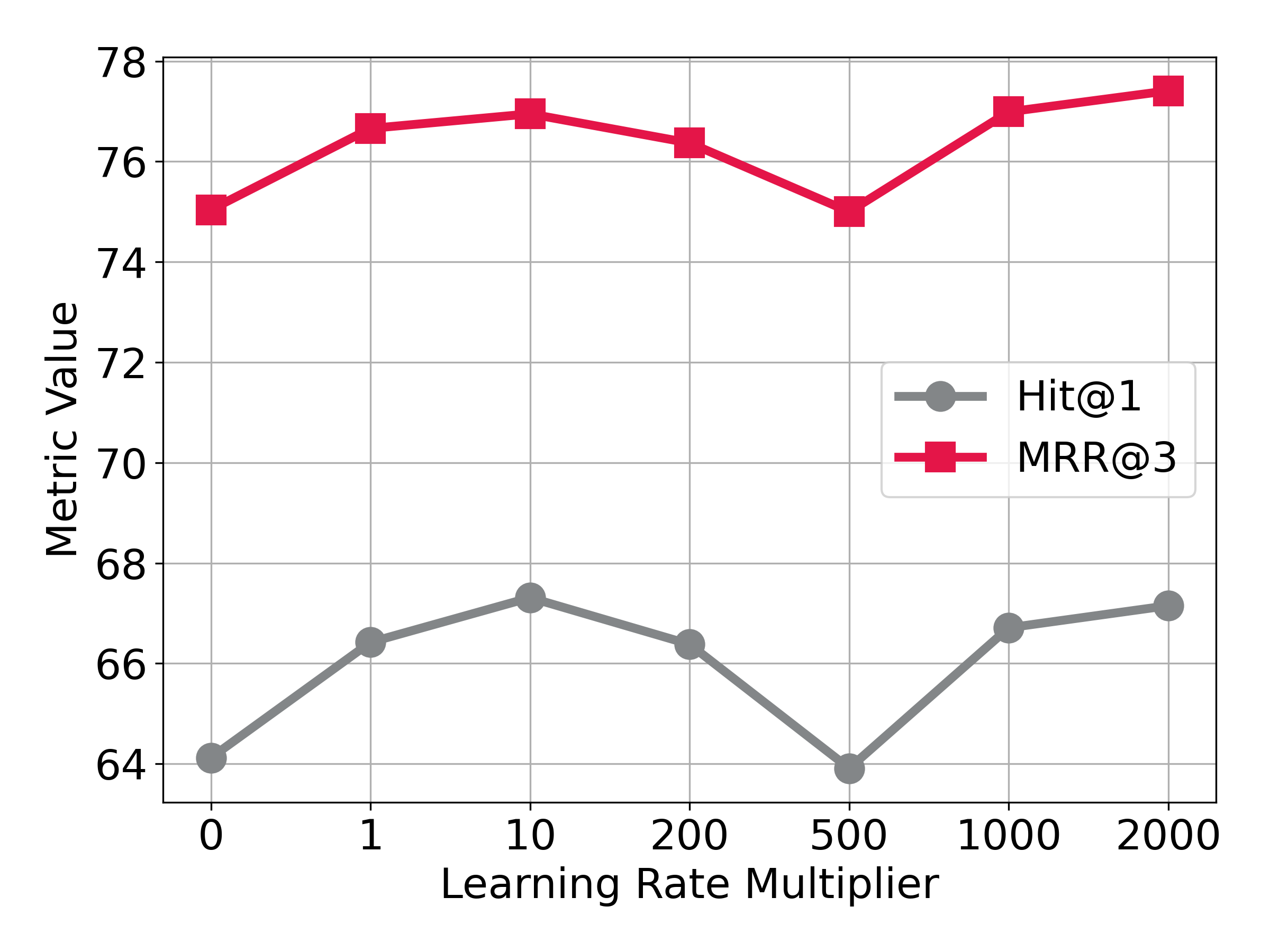}
    \caption{Comparing performance with varying learning rate multiplier ratios on the GeoIND dataset. The learning rate multiplier signifies the ratio of attention matrix learning rate to model parameter learning rate.}
    \label{fig:learning_rate_ratio}
\end{figure}

\subsection{Parameter Sensitivity and Efficiency}

Considering the pivotal impact of the dynamic attention matrix on model performance, we have conducted an additional experiment involving different update speed for model parameters and the attention matrix, which we called asynchronous learning rate updates. The outcomes, as is shown in Figure \ref{fig:learning_rate_ratio}, underline the sub-optimal nature of synchronously updating metrics with model parameters (i.e. ratio=1). Contrarily, we have identified that employing a more extended update step for the attention matrix yields improved results; for instance, setting learning rate ratio at 10 and 2000 for the GeoIND dataset. This trend suggests that the attention matrix carries a weightier importance than general model parameters. Our finding is consistent with similar endeavors focused on adaptively weighted learning \cite{he2017adaptively}. Specifically, within our CGR task, a swifter acquisition of focus by the model on specific geographic chunks reveals to enhanced performance.

Furthermore, in line with our commitment to addressing real-world challenges, it becomes imperative to substantiate the efficacy of the proposed \textit{Geo-Encoder}. To this end, we present an empirical analysis of training and inference times, as detailed in Table \ref{tb:effency_proof}. Evidently, when comparing the results with MGEO-BERT, our training process exhibits a marginal increase in duration due to the incorporation of chunking attention matrix learning and supplementary representation fusion. However, it's noteworthy that our inference times remain remarkably similar, underscoring the effectiveness of our algorithm without causing substantial disparities in computational efficiency. The inference time of all models are acceptable for various industry application scenarios. Moreover, our training time is actually shorter than that of the Augment-Encoder approach \cite{peng-etal-2022-predicate}, demonstrating the effectiveness of multi-task learning rather than geographic component feature concatenation.


\begin{table}[t]
\centering
\begin{adjustbox}{width=0.45\textwidth,center}
\label{tab:correlation}
\begin{tabular}{l|ccc}
\toprule
Model & IndBERT & IndRoBERTa & IndERNIE \\
\midrule
IndBERT & -- & 0.796* & 0.785* \\
IndRoBERTa & 0.796* & -- & 0.932* \\
IndERNIE & 0.785* & 0.932* & -- \\
\midrule
\multicolumn{3}{c}{\phantom{---}} \\
\midrule
Model & TesBERT & TesBERTa & TesERNIE \\
\midrule
TesBERT & -- & 0.819* & 0.604* \\
TesRoBERTa & 0.819* & -- & 0.374\phantom{0} \\
TesERNIE & 0.604* & 0.374\phantom{0} & -- \\
\midrule
\multicolumn{3}{c}{\phantom{---}} \\
\midrule
Model & IndBERT & IndRoBERTa & IndERNIE \\
\midrule
TesBERT & 0.614* & 0.409* & 0.501* \\
TesRoBERTa & 0.713* & 0.634* & 0.672* \\
TesERNIE & 0.253\phantom{0} & 0.035\phantom{0} & 0.175\phantom{0} \\
\bottomrule
\end{tabular}
\end{adjustbox}
\caption{Spearman correlation scores on GeoTES (Tes) and GeoIND (Ind) datasets. Statistically significant results are marked with *, where $\textit{p-value} < 0.05$.}
\label{tb:correlation_spearman}
\end{table}

\begin{figure}[t]
    \centering
    \subfigure[BERT chunk attention weights on GeoIND dataset]{\includegraphics[width=0.43\textwidth]{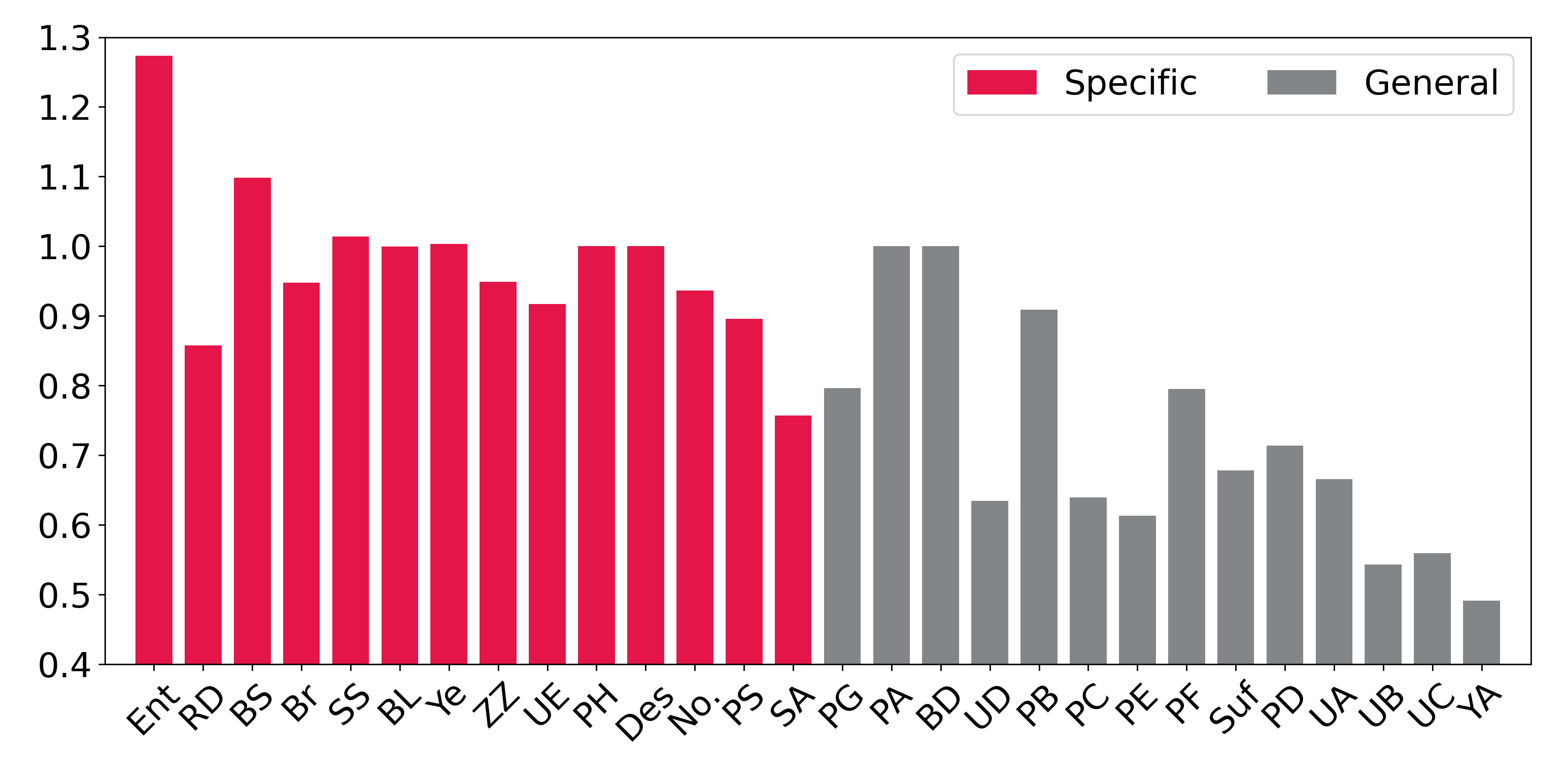}\label{fig:attention_weight_s1}}
    \hspace{0.05\linewidth}
    \subfigure[Statistical distribution of attention matrix]{\includegraphics[width=0.43\textwidth]{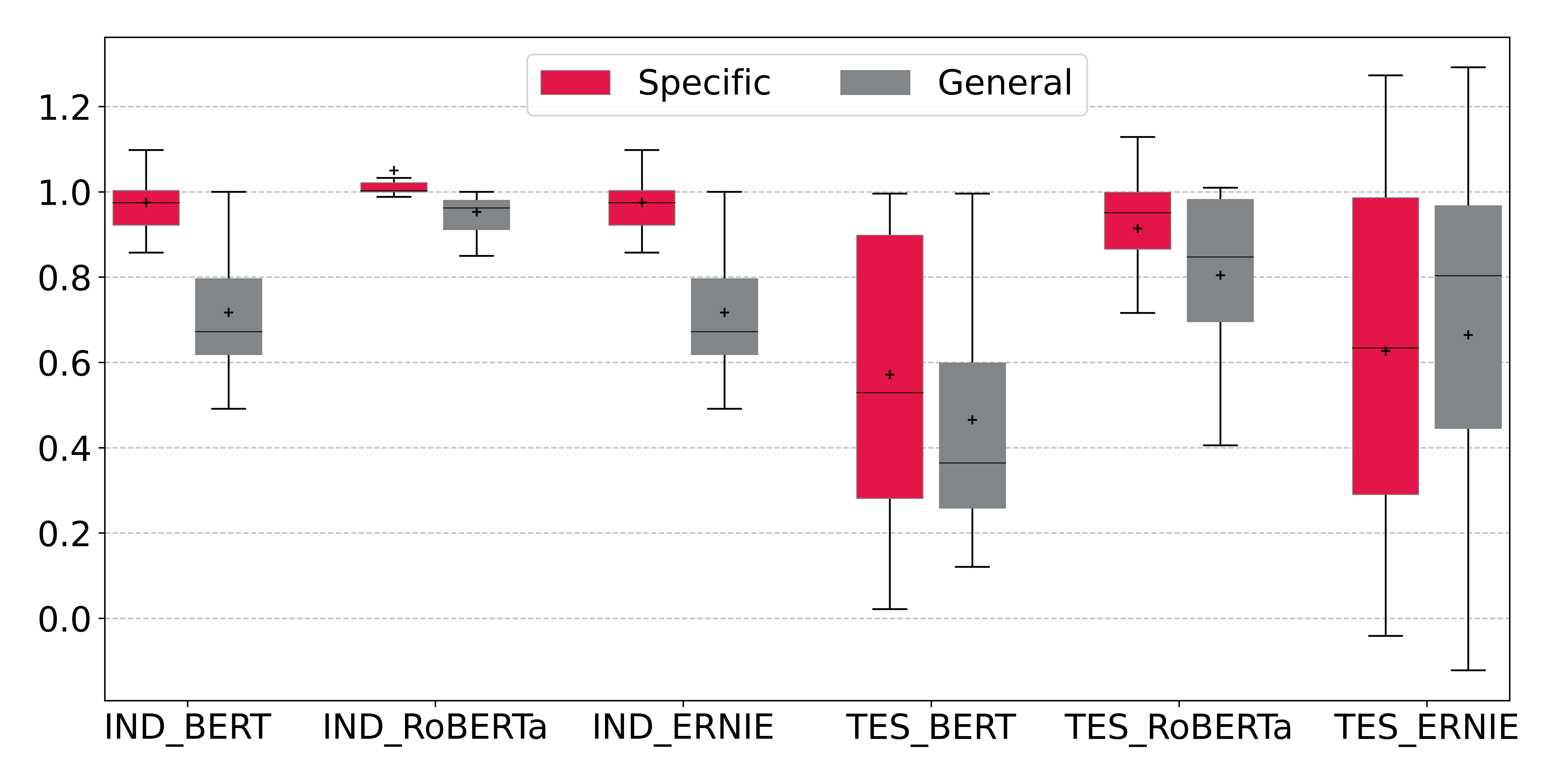}\label{fig:attention_weight_s2}}
    
    \caption{Attention matrix weights visualization. We mark specific chunks as red and  general chunks as grey. Weights of specific chunks are higher than general ones.}
    \label{fig:attention_weight}
\end{figure}

\subsection{Chunking Weight Distribution}

The attention matrix stands as a pivotal element warranting meticulous examination. Consequently, this section delves into an in-depth analysis to discern whether the model demonstrates the capacity to effectively focus on specific chunks as opposed to the more general ones. Employing the MGEO tagging tool, we manually labeled the subsequent categories as specific chunks: bus and subway stations, other administrative districts, branch words, bus and subway lines, house numbers, modifiers, location words, numbers, business district names, encompassing a total of 14 distinct kinds. Conversely, the remaining chunks are classified as general (comprising 15 kinds), such as country, province, city, town, prefix, conjunction, etc.\footnote{All detailed selected chunking labels and its definition can be found in Appendix.}

For enhanced clarity, we manually categorize all chunk types into general and specific classifications, and present the BERT attention matrices in Figure \ref{fig:attention_weight_s1} on GeoIND dataset. Notably, the trend discernible in this figure reveals that specific chunks (red) garner higher weights than general ones (grey). Further, we investigate the tendency across all models and datasets, as depicted in Figure \ref{fig:attention_weight_s2}. The congruence of these outcomes is evident, except for the case of ERNIE on the GeoTES datasets. This discrepancy aligns with the consistent low correlation scores observed between ERNIE and other models, as presented in Table \ref{tb:correlation_spearman}.

Moreover, to probe the consistency across diverse learning processes, we compute spearman correlation coefficients \cite{spearman1961proof} across different datasets. Illustrated in Table \ref{tb:correlation_spearman}, all of these correlation coefficients exhibit positive correlations and most of the results are statistically significant, underscoring uniform learning outcomes in component weights. It is worth noting that, except for the ERNIE model on the GeoTES dataset, the majority of models and datasets exhibit robust correlations, which is obviously evidenced by the high correlation scores. This result aligns with the observation that the ERNIE backbone model attains marginal enhancement, as shown in Table \ref{main_results}. Lastly, models trained on the same datasets yield notably high correlation scores among themselves. For instance, the scores between indBERT and indRoBERTa, and similarly between tesBERT and tesRoBERTa, surpass the 0.78 threshold.

\section{Conclusion}
In this paper, we proposed a novel framework called \textit{Geo-Encoder} for Chinese geographic re-ranking task by deploying multi-task learning module and synchronous update mechanism. The key idea behind \textit{Geo-Encoder} is to encode geographic text using an additional component learning representations from address chunks. 
This approach allows the \textit{Geo-Encoder} to effectively leverage linear-chain characteristic of geographic contexts, which guides the model to capture subtle distinctions among different candidates. 
Moreover, we present an attention matrix that enables the model to automatically learn the significance of geographic chunking components within the representation. To address the varying levels of task complexity, we introduced an asynchronous update mechanism for iterative adjusting the weight matrix of these components. 
This dynamic adjustment facilitates the the focus of model on specific chunks efficiently. 
Extensive experiments demonstrated that our proposed framework leads to significant improvements over several competitive baselines. 
Future work could be incorporating our approach in multi-modal and multi-lingual tasks.

\section{Limitations}
While our work has achieved good performance and shown promising results in enhancing Chinese geographic re-ranking task through incorporation of geographic representations, there are still limitations in our work. Specifically, the Geo-Encoder we have developed exhibits a specificity towards textual data possessing linear-chain or structural characteristics, thereby constraining the method’s applicability primarily to within-domain scenarios. However, we believe that this study is still useful in highlighting the challenges of geographic encoding. Moreover, our approach demonstrates notable effectiveness and efficiency when employed in industrial applications, owing to its minimal augmentation of parameters. 

In the future, we plan to explore the feasibility of collecting multi-modal datasets, which can be potential to provide further insights into incorporating geographic understanding with our proposed framework into CGR task.

\section{Acknowledgement}
Thanks to the anonymous reviewers for their helpful feedback. This work was supported by the China National Natural Science Foundation No. 62202182. The authors gratefully acknowledge financial support from China Scholarship Council. (CSC No. 202206160052). 

\bibliography{anthology,custom}

\clearpage
\appendix

\section{Appendix}
\subsection{Dataset Details}

As previously mentioned, we utilize the MGEO geographic tagging tool\footnote{\url{https://modelscope.cn/models/damo/mgeo_geographic_elements_tagging_chinese_base}.} to thoroughly annotate the provided geographical text. To elaborate further, we present a demonstrative instance in Table \ref{tb:example_geotes}. This example highlights the effectiveness and comprehensive nature of the MGEO in annotating geographical information within the text. 

\subsection{POS Implement}
\label{ax:pos_implement}
We utilize the Jieba tagging tools, which enable the segmentation of all geographical text into meaningful segments. Following this initial breakdown, a rigorous selection process is undertaken, wherein 28 specific parts-of-speech categories are identified as pertinent and aligned with our Geo tagging system. These categories are chosen based on their close relevance to geographical references, thereby ensuring the precision of the tagging process. A comprehensive list of these valid part-of-speech tags is provided in Table \ref{tb:pos_valid}.

In this context, it's important to emphasize that even though manual selection involves a degree of subjectivity, we have maintained consistent tag categories with geographical references to ensure a fair comparison. Additionally, although certain POS tags may not directly pertain to geographic terminology, we have arranged them based on their relative correlations across all POS tags. We have also provided a list of POS tags that are deemed invalid in Table \ref{tb:pos_invalid}, consisting of 24 specific parts-of-speech categories.

Moreover, we compute the fuzzy similarity\footnote{ \url{https://pypi.org/project/fuzzywuzzy/}} between the results of POS tagging and Geo chunking, as shown statistically in Table \ref{tb:sim_pos_geo}.


\begin{table}[htbp]
\centering
\resizebox{0.4\textwidth}{!}{
\begin{tabular}{l|ccc}
\toprule
Set & Avg. Geo & Avg. POS   & Similarity \\
\midrule
\multicolumn{4}{c}
{GeoTES}
\\ \midrule
Train &   5.11 & 10.71 & 80.56 \small{± 7.39}  \\
Dev & 4.69 & 9.47 & 80.46 \small{± 7.35}  \\
Test &  4.66 & 9.41  & 80.60 \small{± 7.41}   \\
\midrule
\multicolumn{4}{c}
{GeoIND}
\\ \midrule
Train & 4.38  & 8.59 & 78.50 \small{± 6.46}      \\
Dev & 4.38  &  8.60 &  79.71 \small{± 6.65}     \\
Test & 4.37 & 8.57 &   79.77 \small{± 6.68}       \\
\bottomrule
\end{tabular}}
\caption{\label{tb:sim_pos_geo} Valid POS categories and their respective definitions, comprising a total of 28 categories.}
\end{table}

\begin{table}[htbp]
\centering
\resizebox{0.35\textwidth}{!}{
\begin{tabular}{l|l|l}
\toprule
{Parameter}   & GeoTES & GeoIND \\
\midrule 
\emph{Learning rate(BERT)} & $5e^{-5}$ & $5e^{-5}$ \\
\emph{Learning rate(RoBERTa)} & $1e^{-5}$ & $1e^{-5}$ \\
\emph{Learning rate(ERNIE)} & $5e^{-5}$ & $5e^{-5}$ \\
\emph{Batch size}    & 32        & 32       \\
\emph{Test Batch size}     & 16      & 16     \\
\emph{Early Stop}  		  & 3    & 3     \\
\emph{Embed\_dim}     & 256      & 256      \\ 
\emph{Optimizer}     & AdamW      & AdamW \\ 
\emph{Attn\_init} & 1.0 & 1.0 \\
\emph{Weight\_decay} & 0.02 & 0.02 \\
\bottomrule
\end{tabular}}
\caption{\label{tb:para_setting} The hyper-parameters of the best results on GeoTES and GeoIND dataset.}
\end{table}

\begin{table}[h]
\centering
\resizebox{0.48\textwidth}{!}{
		\begin{tabular}{l|l}
\toprule
\multirow{2}{*}{{Chunks}} & \multirow{2}{*}{{Definition}} \\
& \\
\midrule
\multicolumn{2}{c}{{General}} \\
\midrule
PA & Country \\
PB & Province \\
PC & City \\
PD & District \\
PE & Township \\
PF & Street \\
PG & Village \\
PH & Administrative Term / Business District \\
PS & Other Administrative Term \\
UA & Door Address: Road xx, No.xx / Lane xx \\
UB & Door Address: Building xx / Area xx \\
UC & Door Address: Building No. xx \\
UD & Door Address: Additional Description \\
\midrule
\multicolumn{2}{c}{{Specific}} \\
\midrule
BS & Bus Station \\
BL & Bus and Subway Route \\
RD & Road, Highway, Furuin Street, Tunnel, Bridge, Overpass \\
Entity & General Name for Point of Interest (POI) \\
Brand & Well-known Brand \\
CategorySuffix & Category Suffix Word \\
Ent & Point of Interest (POI) \\
Br & Brand \\
No. & Number \\
UE & Door Address: East Entrance, South Gate \\
SA & Direction Modifier \\
PH & Administrative Term / Business District \\
Ye & Semantic Connector \\
Des & Descriptor \\
ZZ & Unknown \\
    \bottomrule
\end{tabular}}
	\caption{\label{tb:geo_labels} Translation of Chunking Terms. }
\end{table}

As depicted in Table \ref{tb:sim_pos_geo}, it becomes evident that the average count of Geo chunking units is less than that of POS. Concurrently, a noteworthy inference can be drawn that the chunking outcomes exhibit resemblance. This is supported by the substantial similarity scores (exceeding 78.00) between the results on both datasets.

\subsection{Geo Chunks}

We have compiled a comprehensive table (Table \ref{tb:geo_labels}), that outlines various chunking categories along with their corresponding definitions of Geo chunks. Drawing from our accumulated expertise, we have classified all chunk categories into two distinct groupings: "general" and "specific." 

This categorization is guided by a systematic process that sorts these categories based on their relative degrees of correlation. To elaborate on this process, we strategically designate the first 50\% of the selection as general chunks, while the subsequent 50\% are categorized as specific chunks. By employing this division strategy, we achieve a balanced representation of both general and specific chunk types.

\subsection{Hyper-parameter Setting}

In an effort to support the reproducibility of the \textit{Geo-Encoder} and its demonstrated reasoning performance, we are providing a compilation of the optimal hyperparameters that yielded the best outcomes on two benchmark datasets, as illustrated in Table \ref{tb:para_setting}.

In the process of establishing the baseline, it's important to note that all scores presented in Table \ref{main_results} have undergone training and validation on a consistent hardware platform. Additionally, we are committed to making our baseline code publicly available for reference, which will coincide with the release of our paper.

\begin{table}[t]
\centering
\resizebox{0.35\textwidth}{!}{
		\begin{tabular}{c|l}
  \toprule
{Invalid POS tag} & {Definition} \\
\hline
e & Interjection \\
i & Idiom \\
d & Adverb \\
l & Idiomatic Expression \\
p & Preposition \\
u & Particle \\
y & Modal Particle \\
g & Morpheme \\
x & Non-Morpheme Character \\
vg & Verbal Morpheme \\
vn & Nominal Verb \\
zg & State Morpheme \\
r & Pronoun \\
dg & Adverbial Morpheme \\
tg & Tense Morpheme \\
o & Onomatopoeia \\
uj & Particle \\
ud & Particle \\
nr & Personal Name \\
rg & Modal Particle \\
ul & Tense Particle \\
s & Locative Noun \\
nrt & Personal Name \\
nrfg & Personal Name \\
\bottomrule
\end{tabular}}
	\caption{\label{tb:pos_invalid} Invalid POS categories and their respective definitions, consisting of a total of 24 categories.}
\end{table}

\begin{table}[t]
\centering
\resizebox{0.35\textwidth}{!}{
		\begin{tabular}{c|l}
  \toprule
{Valid POS tag} & {Definition} \\
\hline
nz & Other Proper Noun \\
a & Adjective \\
m & Numeral \\
q & Measure Word \\
t & Time Word \\
mg & Measure Word for Quantity \\
ns & Place Name \\
ng & Noun as Morpheme \\
ag & Adjective as Morpheme \\
f & Locative \\
z & Status Word \\
nt & Organization Name \\
eng & English Word \\
an & Noun \\
mq & Measure Word for Quantity \\
ad & Adverb as Adjective \\
b & Differentiation Word \\
j & Abbreviation \\
n & Noun \\
c & Conjunction \\
uv & Auxiliary Word \\
k & Following Part \\
h & Preceding Part \\
v & Verb \\
uz & Status Word \\
ug & Tense Word \\
df & Differentiation Word \\
yg & Modal Particle \\
\bottomrule
\end{tabular}}
	\caption{\label{tb:pos_valid} Valid POS categories and their respective definitions, comprising a total of 28 categories.}
\end{table}

\CJK{UTF8}{gbsn}
\begin{table*}[t]
\centering
\resizebox{0.75\textwidth}{!}{
		\begin{tabular}{l|l}
			\toprule
			Field  & Content  \\
                \midrule
                \multirow{5}{*}{Query} & 浙江省杭州市人民检察北东院侧广播电视台东门南 \\
                &  South of the East Gate of People's Procuratorate  North \\& East Radio and   Television Station, Hangzhou City, \\& Zhejiang Province. \\ \midrule
                \multirow{5}{*}{Query\_Geo\_Chunks} & 浙江省-prov / 杭州市-city / 人民检察-poi/ 东院-subpoi \\ & / 侧-assist   / 广播电视台-subpoi / 东门-subpoi / 南-assist \\
                &  Zhejiang Province / Hangzhou City /  People's Procuratorate \\ & / East Door / of / Radio and Television Station / East Gate / \\ & South Procuratorate of Hangzhou City, Zhejiang Province. \\
                \midrule
                \multirow{5}{*}{Query\_POS\_Chunks} & 浙江省-ns / 杭州市-ns / 人民-n / 检察-vn / 北东-ns / 院侧-n\\ &/ 广播-vn / 电视台-n / 东门-ns / 南-ns \\
                &  Zhejiang Province / Hangzhou City /  People / Procuratorate /\\ &North East / of / Radio  Television Station / East Gate / South \\ &Procuratorate of Hangzhou City, Zhejiang Province. \\

                \midrule
                \multirow{11}{*}{Candidates} & 浙江省人民北路路旁播州区人民检察院 \\
                & People's Procuratorate of Bozhou District, beside Renmin \\ &North Road,  Zhejiang Province. \\
                & 浙江省人民检察院 \\
                & Zhejiang Provincial People's Procuratorate. \\
                & 浙江省浙江北路136号山东广播电视台 \\
                & Shandong Radio and Television Station, No. 136 Zhejiang \\ &North Road,  Zhejiang Province.  \\
                & 台州路1号杭州市拱墅区人民检察院 \\
                & People's Procuratorate of Gongshu District, Hangzhou City,\\ &No. 1   Taizhou Road. \\ \midrule
                \multirow{14}{*}{Candidates\_Geo\_Chunks} & 浙江省-prov / 人民北路-road / 路旁-assist / \\ & 播州区人民检察院-poi \\
                & Zhejiang Province / Renmin North Road / beside / \\ & People's Procuratorate of Bozhou District. \\
                & 浙江省-prov / 人民检察院-poi \\
                & Zhejiang Province / Provincial People's Procuratorate. \\
                & 浙江省-prov / 浙江北路-road / 136号-roadno / \\ & 山东广播电视台-poi \\
                & Zhejiang Province / Zhejiang North Road / No. 136 \\ & / Shandong Radio and Television Station \\
                & 台州路-road / 1号-roadno / 杭州市-city / \\ & 拱墅区-district / 人民检察院-poi \\
                & Taizhou Road / No. 1 / Hangzhou City / \\ & Gongshu District / People's Procuratorate \\
                \midrule
                \multirow{14}{*}{Candidates\_POS\_Chunks} & 浙江省-ns / 人民-n / 北路-ns / 路旁-s / 播州-ns / \\& 区-n / 人民检察院-nt \\
                & Zhejiang Province / Renmin / North Road / beside / \\ & Bozhou / District / People's Procuratorate. \\
                & 浙江省-ns / 人民检察院-nt \\
                & Zhejiang Province / Provincial People's Procuratorate. \\
                & 浙江省-ns / 浙江-ns / 北路-ns / 136-m / 号-m / \\& 山东-ns / 广播-vn / 电视台-n \\
                & Zhejiang Province / Zhejiang / North Road / 136 / No. \\ & / Shandong / Radio / Television Station \\
                & 台州-ns / 路-n / 1-m / 号-m / 杭州市-ns / 拱墅区-ns / \\& 人民检察院-nt \\
                & Taizhou / Road / 1 / No. / Hangzhou City / \\ & Gongshu District / People's Procuratorate \\

    \bottomrule
\end{tabular}}
	\caption{\label{tb:example_geotes} A representative illustration sourced from the GeoTES dataset is provided. We are showcasing a subset of potential options in this context. The English was meticulously translated, as this information isn't inherently present in our initial dataset. }
\end{table*}

\end{document}